\documentclass[runningheads]{llncs}
\usepackage{graphicx}
\usepackage[usenames,dvipsnames]{color} %
\usepackage{amsmath,epsfig,amssymb}
\usepackage{svg}
\newcommand{\smallsqueezeup}{\vspace{-2.5mm}}
\newcommand{\squeezeup}{\vspace{-5.0mm}}
\newcommand{\bigsqueezeup}{\vspace{-8.0mm}}

\begin{document}
\title{Weak Segmentation-Guided GAN for realistic color edition}
%
%
\author{Vincent Auriau\inst{1} \and
Emmanuel Malherbe\inst{1,2} \and
Matthieu Perrot\inst{1}}
\authorrunning{V. Auriau et al.}
%
\institute{L'Oréal AI Research, France \email{\{vincent.auriau,matthieu.perrot\}@loreal.com} \and
Artefact Research Center, France
\email{emmanuel.malherbe@artefact.com\footnote{part of the job was done while employed at L'Oréal}}\\
}

\maketitle              
\bigsqueezeup
\begin{figure}[h!]
\centering
\includegraphics[width=0.9\linewidth]{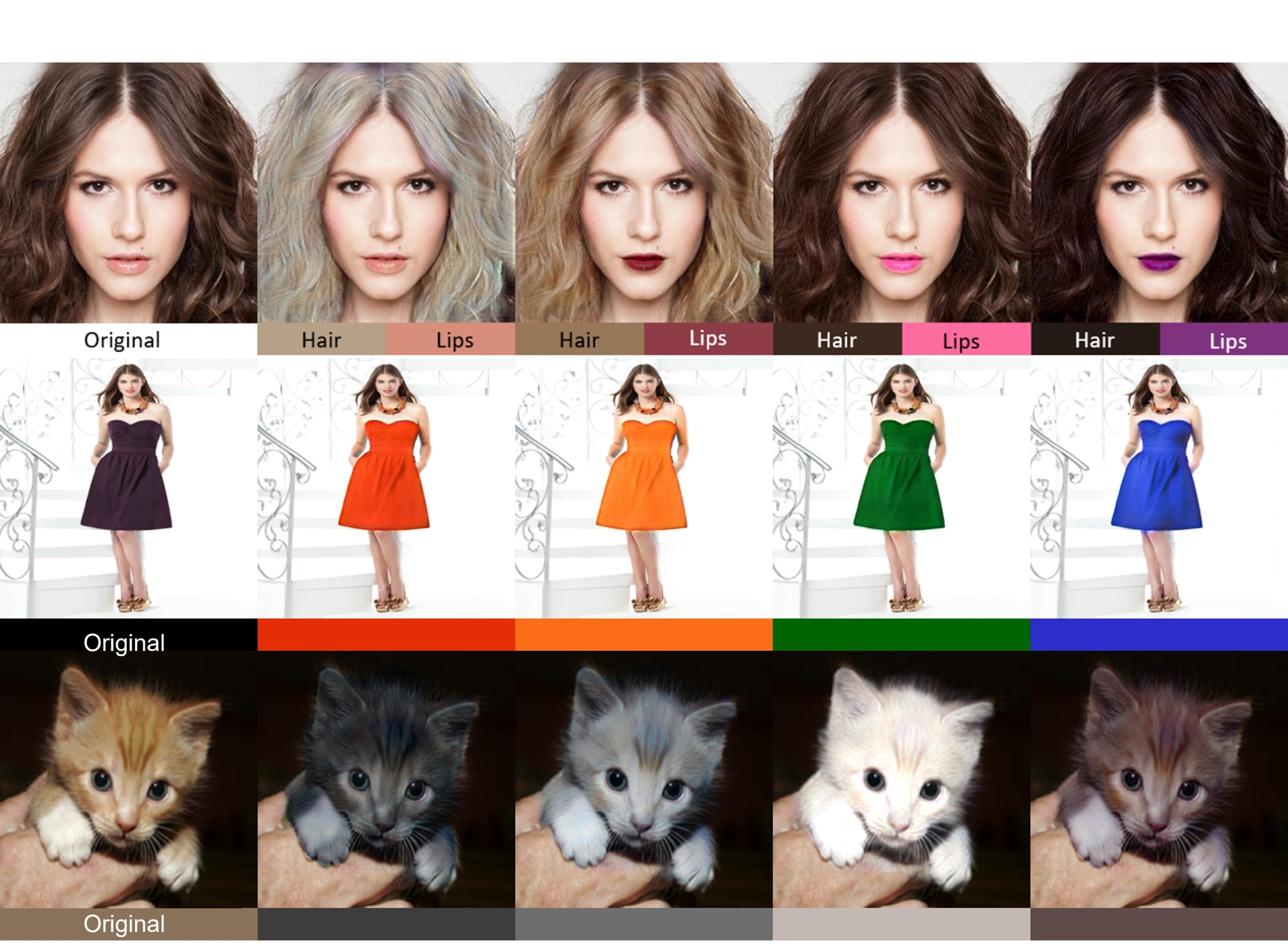}

\caption{Example results of our framework on three different use cases: coloring hair and lips, clothing and cats and dogs.}
\label{fig:entry_stack}
\end{figure}
\bigsqueezeup
\smallsqueezeup

\begin{abstract}
 Editing the color of images in a realistic way finds many applications such as changing the perception of an image, data augmentation or film post processing. The design of an automatic tool is a complex and long addressed challenge. In particular, two properties are difficult to meet altogether: generating realistic results without artifacts and the possibility to precisely choose the future color. Conventional methods using segmentation and histogram matching maximize the controllability but also introduce a lack of realism or are complex to automate. On the contrary, GANs that specialize in realism are difficult to control. To overcome these challenges, we propose a novel GAN architecture leveraging any differentiable segmentation model. We demonstrate the genericness of our framework that presents state of the art results on different use cases. It generates images that look realistic while offering a precise color control.

\end{abstract}
\section{Introduction}

\label{sec:intro}
Color edition control is the process of modifying an object's color in a graphical representation. A user can control the edition by specifying desired colors.
It is a great tool for images post processing. By altering the color palette, it is possible to modify the global atmosphere or perception of the image \cite{10.2312:COMPAESTH:COMPAESTH12:089-097}. One might want to be more precise and only adjust the color of a single area: a lighter blue sky in a landscape photograph or changing an actor's hair color in a movie. Simulating several color variations of the same car or garment can help to choose the right color among the millions of possibilities before any material realization. It can also be used to help customers select the right make up or hair coloration they desire. While physical renderers handle colors very well \cite{sota_neural_rendering}, it is still a real challenge to work with existing photographic pictures.
Color transfer \cite{overview_style} and colorization \cite{7900208} are two other applications that can be seen as specific cases of color edition. Most recent methodologies use neural networks and particularly GANs \cite{chromagan}, however they still meet challenges with some non satisfying results.

Our goal is to realistically and precisely change an object's color in an image. We consider the general color perception of an object in the image, characterizing the object by a single color. In order to keep the human control simple, interaction of light with the object's geometry or local color variations coming from the textures are to be handled automatically. 

To better define the objective, we enumerate four properties to evaluate the performances of a color edition model:

\begin{itemize}
    \item [(i)] \textbf{Colorization accuracy:} The perceived resulting color of the object is as close as possible as the desired one
    \item[(ii)] \textbf{Realism:} The generated image is realistic to the human eye
    \item[(iii)] \textbf{Stability:} The geometry of the considered object and local color variations are kept unchanged
    \item [(iv)] \textbf{Restraint:} Objects and areas that should not be modified, or "background",  are left untouched
\end{itemize}

We describe our robust method for color edition that overcomes the two main difficulties of existing techniques: producing realistic results while reaching a precise color control. Our main contributions are:
\begin{itemize}
    \item A generic model architecture and training strategy for editing human perceived color while ensuring realism, stability and constraint
    \item A versatile object-aware color edition tool based on segmentation that addresses various use-cases and applications: color edition, color transfer and colorization
    \item Experimental comparison on various datasets and use cases that show how our approach achieves the best trade-off compared to the state-of-the art models
\end{itemize}

\section{Related Work}

Color manipulation in images is a long-addressed problem. Different semi-automatic or fully automatic methods have been developed, with different advantages and drawbacks. We specifically focus on color edition on the core of this paper, however inspirational technologies have been developed for color transfer or colorization.

\textbf{Color Transfer} consists of extracting the right color features from a reference image, and to transfer them to the content image. In \cite{Reinhard2001ColorTB}, a simple but effective function for color transfer is given. The color reference image $X$ and the content image $\tilde{X}$ are described respectively by $(\mu, \sigma)$ and $(\tilde{\mu}, \tilde{\sigma})$, the mean and standard deviation of the color distributions of each image in the L*a*b* space. The color transfer function is then defined as:
\begin{equation}
    g(\tilde{X}_{i, j}) = \mu + \frac{\sigma}{\tilde{\sigma}}(\tilde{X}_{i, j}-\tilde{\mu})
    \label{eq:rein_transfo}
\end{equation}
More complex transformation manipulate the whole color histogram \cite{10.5555/2381219.2381235}\cite{DBLP:journals/corr/abs-1907-06740}. It enables modifications of the color palettes that would not be possible with a simple linear translation. In return, the needed computation is much greater.
Most state of the art are today produced by deep learning. In \cite{DBLP:journals/corr/abs-1710-00756} a neural color transfer method between images is proposed. It uses a trained VGG-19 to match semantically meaningful entities between the reference and content images, before applying Equation \ref{eq:rein_transfo}.
While having an image to transfer the color from is of great help, it might be long or complicated to find the very example image that perfectly fits the user's intention. It also limits creativity, forcing the use of already existing color palettes.

Convolution neural network have also shown the first good results on automatic \textbf{Gray Image Colorization} \cite{he2018attgan}. The use of conditional GAN with the pix2pix model \cite{DBLP:journals/corr/IsolaZZE16} have led to more realistic results. These methods do not allow any color controlability as the resulting color cannot be decided by the user. Such control is introduced in \cite{https://doi.org/10.48550/arxiv.1612.00835} with colored drawn strokes. No measure of the color precision is developed and background is often impacted, creating undesired artifacts.

In \textbf{Color Edition Control}, a user can modify at will the colors of an object in an image. Pixels to be edited can be selected using a segmentation model \cite{PAL19931277}, which was more recently performed using neural segmentation \cite{levinshtein2018realtime,DBLP:journals/corr/abs-1907-06740}. The desired color is defined by the user in the form of a histogram or color look up tables. Respectively histogram matching algorithms and interpolation are then applied to modify the selected pixels. While providing good results, these methods suffer in terms of realism, particularly around the edges of the computed mask where artifacts tend to appear. Moreover the color formats are complicated to manipulate for a user.

Another strategy that overcomes those limitations is to use a conditional GAN architecture \cite{DBLP:journals/corr/MirzaO14} such as the one of StarGAN \cite{DBLP:journals/corr/abs-1711-09020}, with the colors as condition variables. CAGAN \cite{DBLP:journals/corr/abs-2008-10298} proposes to automatically extract colors and let the generator indirectly focus on the right area without further explicit supervision. The major persisting challenges are generated images color precision and the inherent modifications of background pixels, that create inconsistencies. To overcome this issue, MagGAN \cite{DBLP:journals/corr/abs-2010-01424} suggests to feed the model with manually labelled masks to improve the focus on a specific zone. However it is not designed for continuous color space and requires mask labels as input.

In this paper, we aim at bridging the different approaches presented above, GANs for realism and segmentation for object selection. It leads us to create an accurate, controllable and versatile coloring tool for edition.

\section{Proposed model}

\label{section:proposed_framework}

Our approach consists of Generator $G$ that produces the transformed image $\hat{X} = G(X, c)$ from the input image $X$, with the considered object colorized as described by the color vector $c$. We also define $S$, a generic and differentiable segmentation model such that $S(X)$ is the probability for each pixel to represent the object: $S(X) \in[0; 1]^{W\times H}$ where $S(X)_{i, j}=\mathbb{P}(x_{i, j}\in object)$. In practice we use a neural network for $S$ that is trained separately from the rest of the framework and frozen, but it can be generalized to any differentiable function.

\begin{figure}
    \centering
      
  \includegraphics[width=1.0\linewidth]{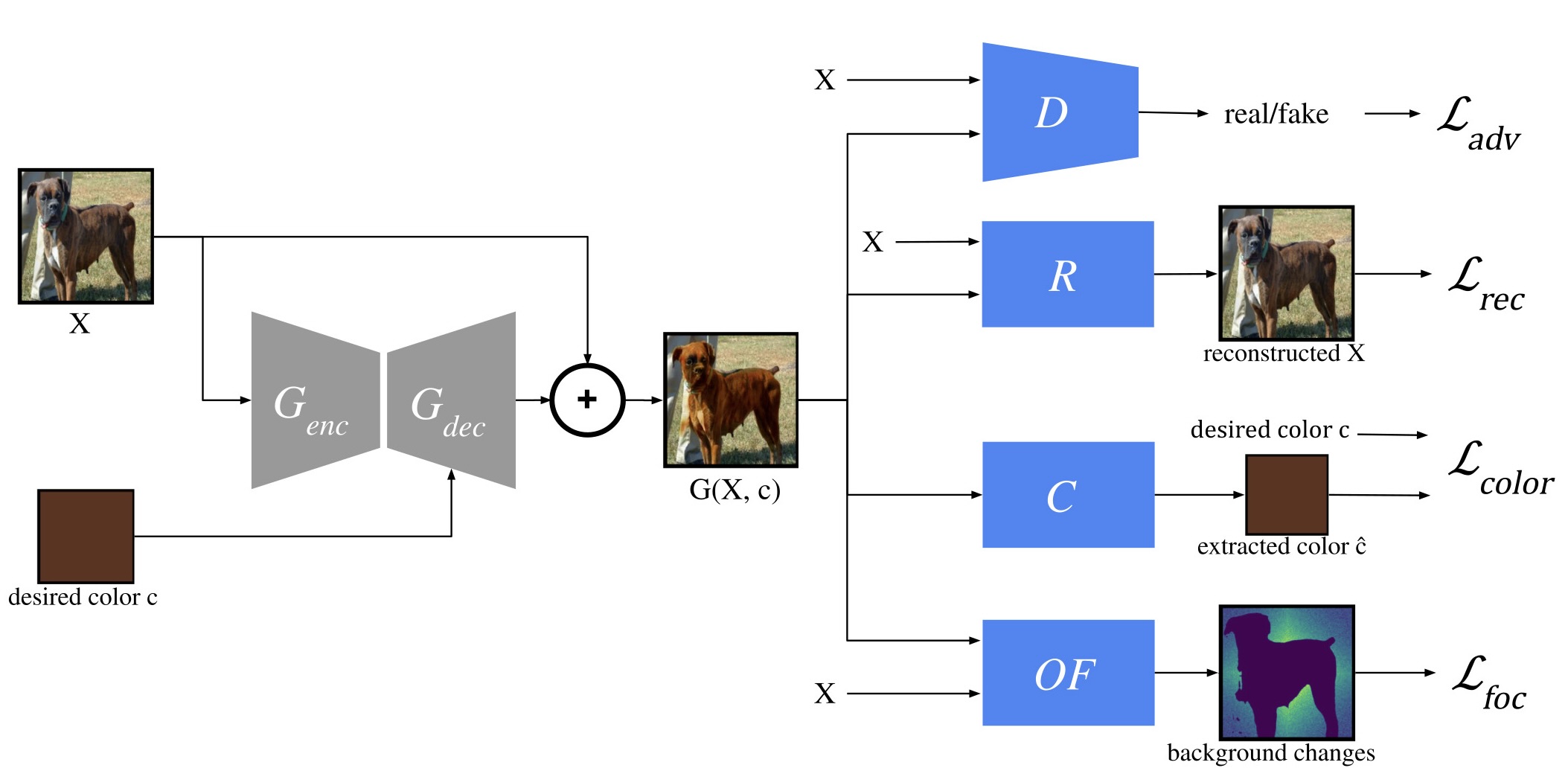}
    
    \caption{General organization of our framework. Generated images are evaluated through the four defined modules that design $G$'s training: $C$, $D$, $OF$ and $R$. $G$ actually predicts a difference that is applied to the input image.}
    \label{fig:losses_computation}
\end{figure}

$G$ is a convolutional neural network with an encoder-decoder architecture. The decoder takes the image features created by the encoder and the desired color as inputs and generates the colorized image. $G$ is trained to optimize four modules that ensure the required properties: $C$ for colorization accuracy, $D$ for realism, $R$ for stability and finally $OF$ for restraint. The general organisation is illustrated Figure \ref{fig:losses_computation}, and the modules described in the following sections.

\squeezeup

\smallsqueezeup
\subsection{Color Estimation module $C$}
\label{subsection:controlling_color}

\begin{figure}
\centering
    
  \includegraphics[width=0.85\linewidth]{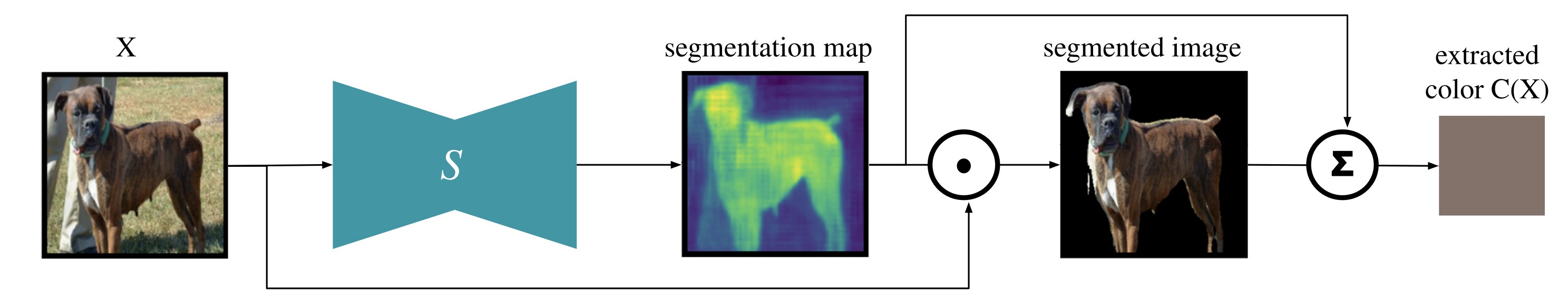}
    
  \caption{Color estimation module $C$: from the segmentation map given by the segmentation model $S$, the weighted mean $C(X)$ is computed to represent the color of the object.}
  \label{fig:color_est}

\end{figure}

We need a robust color estimation function that should be differentiable for later optimization purposes.
We describe an object's color as an average $RGB$ value over its representing pixels. This estimator characterizes well the general color of a uniformly colored object and can be easily represented and understood by a user. The weighted mean color of an object $C(X)$ is then:
\begin{equation}
    C(X) = \mathbb{E}_{object}[X] = \frac{\sum_{i, j}{S(X)_{i, j} * x_{i, j}}}{\sum_{i, j}{S(X)_{i, j}}}
    \label{'color_computation'}
\end{equation}
 with $x_{i, j}$ the color vector at pixel (i, j). It is illustrated in Figure \ref{fig:color_est}.
 
\smallsqueezeup
This color estimator is used to estimate the color of our object in the generated image $\hat{X}$ and compare it to the desired color $c$. The color loss is a $L_2$ loss computed in the $CIEL^*a^*b^*_{76}$ color space as it better matches human perception \cite{DBLP:journals/corr/abs-2008-10298,Reinhard_colourspaces}.
\smallsqueezeup
\begin{align}
    \label{eq:color_loss}
    \mathcal{L}_{color} = || C(\hat{X})-c ||_{L^*a^*b^*_{76}}
\end{align}

Thanks to the use of $S$, we are sure to extract and therefore control the color of the object itself.

\subsection{Discriminator module $D$}
The GAN structure ensures that the main coloring model $G$ produces realistic results by adversarially training a discriminator neural network, $D$. It tries to distinguish real images from the one edited by $G$. Our adversarial loss is the standard Wasserstein GAN loss with gradient penalty \cite{arjovsky2017wasserstein}:

\begin{align}
    \label{eq:adv_loss}
    \mathcal{L}_{adv} = D(\hat{X}) - D(X) + \lambda_{gp} \nabla_{\Tilde{X}}D(\Tilde{X})
\end{align} with $\Tilde{X}$ randomly obtained for each batch by interpolating between real images $X$ and generated images $\hat{X}$.

\subsection{Reconstruction module $R$}

\begin{figure}
 \centering
 
  \includegraphics[width=0.7\linewidth]{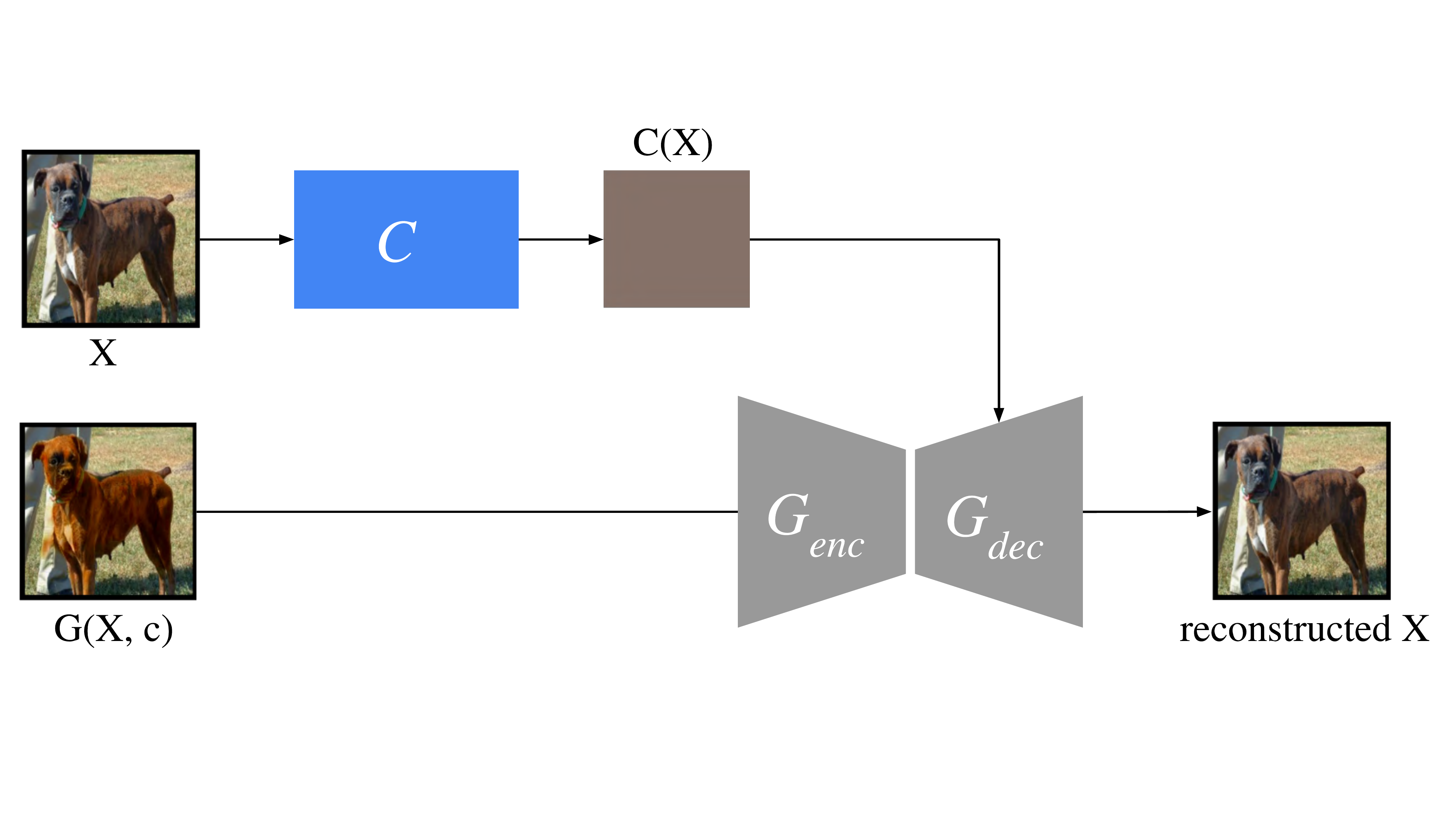}

  \caption{Reconstruction module $R$: The edition model $G$ is asked to reconstruct an original image from one of its own edition.}
  \label{fig:rec_module}

\end{figure}

In order to ensure stability, we use the popular cycle consistency loss introduced in CycleGAN \cite{zhu2020unpaired} for image to image translation. The model is trained to be able to recreate the original image from its own generation. It ensures that $G$ maintains the objects geometry and that no pixel or information is lost through the process. The reconstruction loss is defined as:
\begin{equation}
\label{eq:rec_loss}
\mathcal{L}_{rec} = ||X - G(G(X, c), C(X))||^2    
\end{equation}
The reconstruction module is illustrated in Figure \ref{fig:rec_module}.

\bigsqueezeup
\subsection{Object Focus module $OF$}
\begin{figure}
 \centering

  \includegraphics[width=0.7\linewidth]{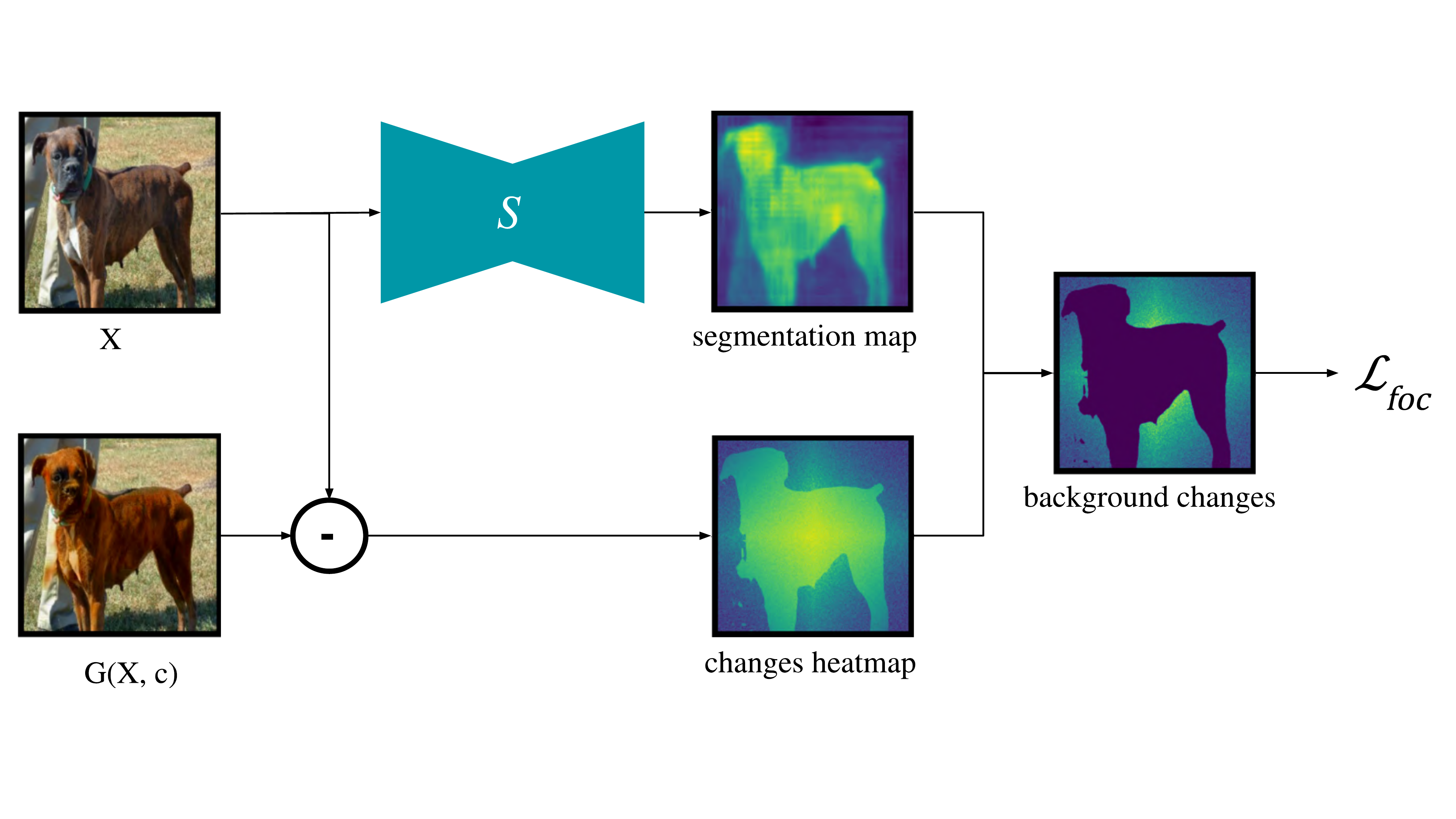}

  \caption{Object focus module $OF$: Generated images are compared to the original ones and all pixels modified outside the segmentation maps are penalized. This helps to keep a background consistency between original and generated images.}
  \label{fig:object_focus}

\end{figure}

With our frozen segmentation model $S$, we can also ensure restraint, meaning that $G$ keeps background pixels unchanged. We only want pixels that relate to the object's color change to be edited by the model. These background alterations is a common artifact produced by GAN based approaches. The Object Focus loss $\mathcal{L}_{foc}$ penalizes any change operated outside of the segmentation mask representing our object:

\begin{equation}
    \label{eq:object_focus_loss}
    \mathcal{L}_{foc} = ||(X - \hat{X})) \odot (\textbf{1}-S(X))||^2
\end{equation}
with $\textbf{1}$ the matrix of same shape as $S(X)$ and filled with ones and $\odot$ the element wise product on the R, G and B channels. The resulting object focus module is illustrated in Figure \ref{fig:object_focus}.
\smallsqueezeup

\subsection{General objective}
We summarize the final objective to train $G$ and $D$ as follows:

\begin{equation}
    \mathcal{L}_G = \mathcal{L}_{adv} + \lambda_{color}\mathcal{L}_{color} + \lambda_{foc}\mathcal{L}_{foc} + \lambda_{rec}\mathcal{L}_{rec}
\end{equation}
\begin{equation}
    \mathcal{L}_D = -\mathcal{L}_{adv}
\end{equation}
where the different $\lambda$ are used to set up and control the importance of the different losses in the final objective. The general workflow of the model is illustrated in Figure \ref{fig:losses_computation}.
Models $G$ and $D$ are trained by this process, $S$ is trained beforehand separately.

We have presented our framework to edit a single object's color. It can be easily extended to any number of objects. One only needs to sum the losses over the different objects. 

\section{Experiments}

\subsection{Datasets}
We have tested our framework on three different use cases supported by open-source datasets designed to train the segmentation model. First, we used CelebA-HQ dataset \cite{CelebAMask-HQ} to train our hair and lips segmentation network. It was then completed with the FFHQ dataset \cite{DBLP:journals/corr/abs-1812-04948}  to improve the data diversity.
Secondly, the cat and dogs dataset \cite{catsanddogs} was used for coloring pets. For this use case, we used an already trained segmentation network from \cite{kerasdeeplabv3plus}. We plugged this model to our framework that works without further need of adaptation. Finally, the clothing dataset \cite{clothing2} was used to train the segmentation model as well as the color edition model for the garnments use-case. Example of results can be found in Figure \ref{fig:entry_stack}.
The results presented in this part are all obtained using the first datasets. Additional material and examples are shared in the appendix \footnote{\label{footnote_appendix}Appendix can be found at: \url{https://github.com/VincentAuriau/wsg-gan-color-edition/blob/main/pdf\_files/appendix.pdf}}.

\subsection{Implementation}
For the following experiments, the parameters described in this part were used, unless it is specifically specified otherwise. The models were trained on $256\times256$ pixels images resized by bicubic interpolation. For the trained segmentation models, we used a U-Net segmentation network \cite{DBLP:journals/corr/RonnebergerFB15}. The generator and discriminator architectures can be found in the Appendix\footnotemark[\value{footnote}]. For losses weighting, we used the parameters: $\lambda_{rec}=1$, $\lambda_{cond}=80$ , $\lambda_{foc}=600$ and $\lambda_{gp}=6$. We trained both networks with the Adam optimizer with learning rates $5\times10^5$ for the generator and $10^5$ for the discriminator.

\subsection{Evaluation Protocol}
In order to compare our framework with other approaches, we use quantitative metrics computed with the following protocol:
We select a random subset $\mathcal{X}_1$ composed of 10k images that we keep for testing. We select a second subset $\mathcal{X}_2$, of same size as $\mathcal{X}_1$. We generate $G(X_1^i, C(X_2^i))$ for each $(X_1^i, X_2^i)\in(\mathcal{X}_1, \mathcal{X}_2)$ to build the dataset $\mathcal{X}_{1 \rightarrow 2}$. 

\paragraph{Metrics}
For realism quantification, we use the Fréchet Inception Distance, or FID \cite{DBLP:journals/corr/abs-2106-08575}. We compare our generated images $\mathcal{X}_{1\rightarrow2}$ with $\mathcal{X}_2$:
$FID_{1\rightarrow2}=FID(\mathcal{X}_{1\rightarrow2}, \mathcal{X}_2)$. For color precision estimation, we use $L_{1\rightarrow2}^{color} = \frac{1}{n}\sum_{i=1}^n||C(X_{1\rightarrow2}^i)-C(X_2^i)||_{L^*a^*b^*_{76}}$.
We also compute a modified version of $\mathcal{L}_{foc}$ to quantify how well the different models perform on the task of focusing on the right pixels. $L_{bg}^{avg}$ is defined as:
\begin{equation}
    L_{bg}^{avg} = \frac{1}{3n}\sum_{i=1}^n \frac{||(X_{1\rightarrow2}^i - X^i_1) \odot (\textbf{1} - M_1^i)||_1}{||\textbf{1} - M_1^i||_1}
    \label{eq:L_mask_avg}
\end{equation}
with $(X_1^i)_n$ being all the images of $X_1$, $M^i_1$ the manually annotated mask corresponding to $X_1^i$,  $X_{1\rightarrow2}^i$ the corresponding generated image and $\textbf{1}$ the matrix filled with ones.
$L_{bg}^{avg}$ represents the average modification of a background pixel's channel. The $\mathcal{L}_1$ norm is taken to better express human perception in $RGB$ \cite{7324174}.

\paragraph{User test}
We have performed a user study for human quantitative evaluation of the different approaches. We have asked five annotators which was the most realistic edition among the one generated by the different approaches. We have repeated the experience with 30 random triplets of editions, totalling 150 annotations.

\subsection{Comparison with other approaches}
We compare our framework with other approaches that aim at coloring objects in images. 
The first approach, alpha blending \cite{levinshtein2018realtime}\cite{10.5555/2381219.2381235} uses segmentation paired with RGB shift. For fair comparison, the area to edit is selected using $S$, then Equation \ref{eq:rein_transfo} is applied for the average color of segmented pixels to match the desired one.

The second one, CAGAN \cite{DBLP:journals/corr/abs-2008-10298}, is a straightforward adaptation of StarGAN \cite{DBLP:journals/corr/abs-1711-09020} for color control. A third model is trained besides $D$ for color extraction. It uses labels, without any guidance on the object. For fair comparison we extracted color labels using $S$. This third model induces noise in the global color precision of the editions.

Finally, we also benchmark our model with one of the most performing GAN, StyleGAN \cite{DBLP:journals/corr/abs-1912-04958}. As it is not built specifically for coloring, we developed a straight forward strategy on the style space \cite{DBLP:journals/corr/abs-2011-12799} which is detailed in the Appendix \footnotemark[\value{footnote}]. This simple adaptation enables to perform color edition using StyleGAN. More complex adaptation could be designed, but it is beyond the scope of this paper. 

\begin{figure}
  \centering
  \includegraphics[width=1.0\linewidth]{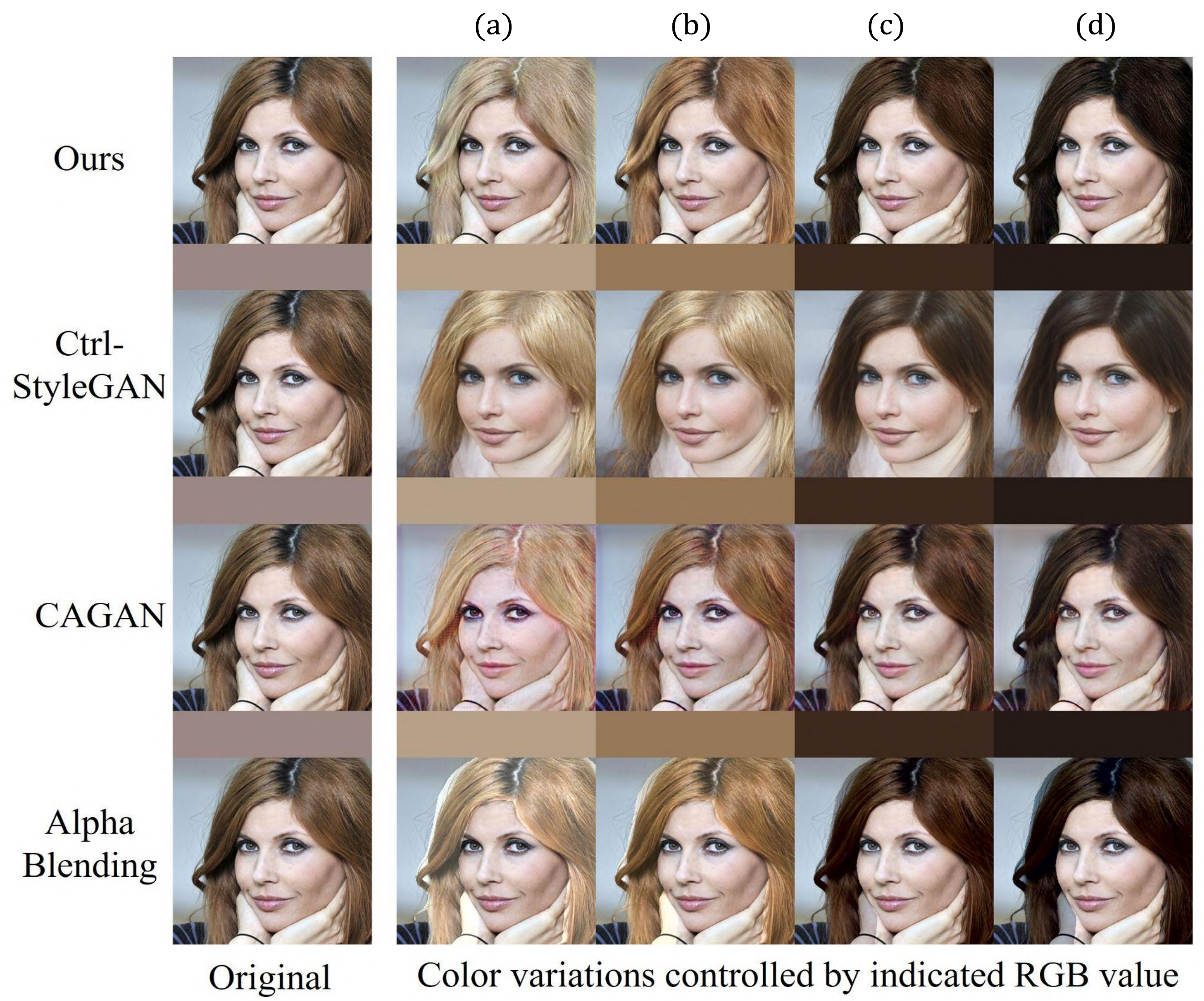}

  \caption{Comparison of different methods for hair coloring. The desired hair color is illustrated on each column between rows.}
  \label{fig:hair_methods_comparison}
\end{figure}

\bigsqueezeup

\begin{table}

\centering
\caption{Quantitative results of the color edition frameworks. We show that our approach outperforms alpha blending in terms of $FID$ and other approaches on all metrics. Our model also outperforms others on the users votes. For the first three metrics, lower is better. Best results are in bold and second best in italic.}
\label{tab:approaches_metrics}
\begin{tabular}{|l|c|c|c|c|}
\hline
Model & $L_{bg}^{avg}\times10^3$ & $FID_{1\rightarrow2}$ & $L_{1\rightarrow2}^{color}$ & User \% Most Realistic \\
\hline
Ours & \textit{0.92} & \textbf{0.050} & \textit{5.03} & \textbf{61\%} \\
Ctrl- StyleGAN & 9.51 & 0.670 & 17.2 & 0\% \\
CAGAN & 121 & \textit{0.096} & 11.3 & 0\% \\
Alpha Blending & \textbf{0.00} & 1.301 & \textbf{0.00} & \textit{34\%}\\
\hline

\end{tabular}
\end{table}

\paragraph{Qualitative evaluation}
Figure \ref{fig:hair_methods_comparison} shows examples of results for hair colorization. We observed the main defaults of compared approaches. Alpha Blending creates artifacts on the edges of the segmented mask (see (a)). CAGAN tends to edit pixels outside hair pixels. It is particularly obvious on (a) with the face color that is different and on (d) where the background is affected. Finally, StyleGAN which is not built for image-to-image translation changes the faces attributes, the projection failing at perfectly reproducing the original image.
\paragraph{Quantitative evaluation}Table \ref{tab:approaches_metrics} shows a metric analysis of the approaches. We observed that our model is a good balance between color precision and realism. Our objective was to compensate alpha blending’s lack of realism using GANs. However, typical GAN approaches come at a cost, particularly on color precision. In addition, it is more difficult to control the outcome of the GAN. It can be seen with $L_{bg}^{avg}$ for CAGAN and ctrl-StyleGAN, showing that many pixels are modified whereas they should not. We observe that our model outperforms other GAN approaches on these topics.

Our approach obtained the majority of users votes and particularly outperforms the other GAN-based approach, CAGAN. Compared to Alpha Blending, our method seems to have better results particularly with the lack of textures in images with darker colored objects.

\subsection{Ablation Study}
We have also analysed the behaviour of the different modules within our framework. We have trained our model and  computed the metrics with and without the reconstruction and object focus losses. The results can be found in Table \ref{tab:ablation_study}.
In order to emphasize the importance of $S$, we also train $GTmasks$, an equivalent of our framework with losses computed with ground truth masks instead of $S$'s predictions. We test this strategy for Object Focus, $GTmask_{OF}$, and combined with Color Estimation, $GTmask_{C+OF}$. $GTmask_{C+OF}$ is thus optimized without a fully differentiable $C(X)$. 

\begin{table}
\centering
\caption{Ablation study of our proposed model. The Object Focus loss improves realism and background preservation while reconstruction positively affects all metrics. For all metrics, lower is better. Best results are in bold, identity excluded.}
\label{tab:ablation_study}
\begin{tabular}{|l|c|c|c|}
\hline
Model & $L_{bg}^{avg}\times10^3$ & $FID_{1\rightarrow2}$ & $L_{1\rightarrow2}^{color}$ \\
\hline
\textit{Identity} & \textit{0.00} & \textit{0.045} & \textit{25.2} \\
$\lambda_{foc}=0, \lambda_{rec}=0$ & 8.38 & {0.051} & \textbf{4.46} \\
$\lambda_{foc}=0, \lambda_{rec}=1$ & 8.00 & 0.060 & {4.80} \\
$\lambda_{foc}=600, \lambda_{rec}=0$ & {1.01} & 0.056 & 5.99 \\
Ours & \textbf{0.92} & \textbf{0.050} & 5.03 \\
$GTmask_{C+OF}$ & 3.40 & 0.082 & 8.51 \\
$GTmask_{OF}$ & 1.79 & 0.052 & 5.20 \\
\hline
\end{tabular}
\end{table}

We observe that optimising the object focus loss fulfills the objective of limiting the modification of pixels outside of the colored object. Combined with reconstruction loss, it further improves all metrics. Finally, the use of a differentiable $S$ function considerably improves the results.

\section{Conclusion}

We presented our framework for generic objects colorizing. Our new object-aware colorization model based on segmentation, introduces specific losses to improve the realism and the color precision of the results. It can be used on any type of image of objects and only needs a pretrained and differentiable segmentation model to work. We have shown state of the art results on several datasets, with the best balance between realism of generated images, accurate color of transformed object as well as maintained geometry and background. Our framework can be adapted for different applications such as color edition,  color transfer or colorization, as shown through the experiments on several use-cases. As future work, we will focus on spatial distribution of the color. In order to let the user manipulate more than one general color of the object while keeping the simplicity of use.

%
%
%
\bibliographystyle{splncs04}
\bibliography{bib}

\clearpage
\section{Appendix}
\subsection{Implementation Details}
\label{annex:implementation_details}
\begin{figure}
    \centering
    \includegraphics[width=1.0\linewidth]{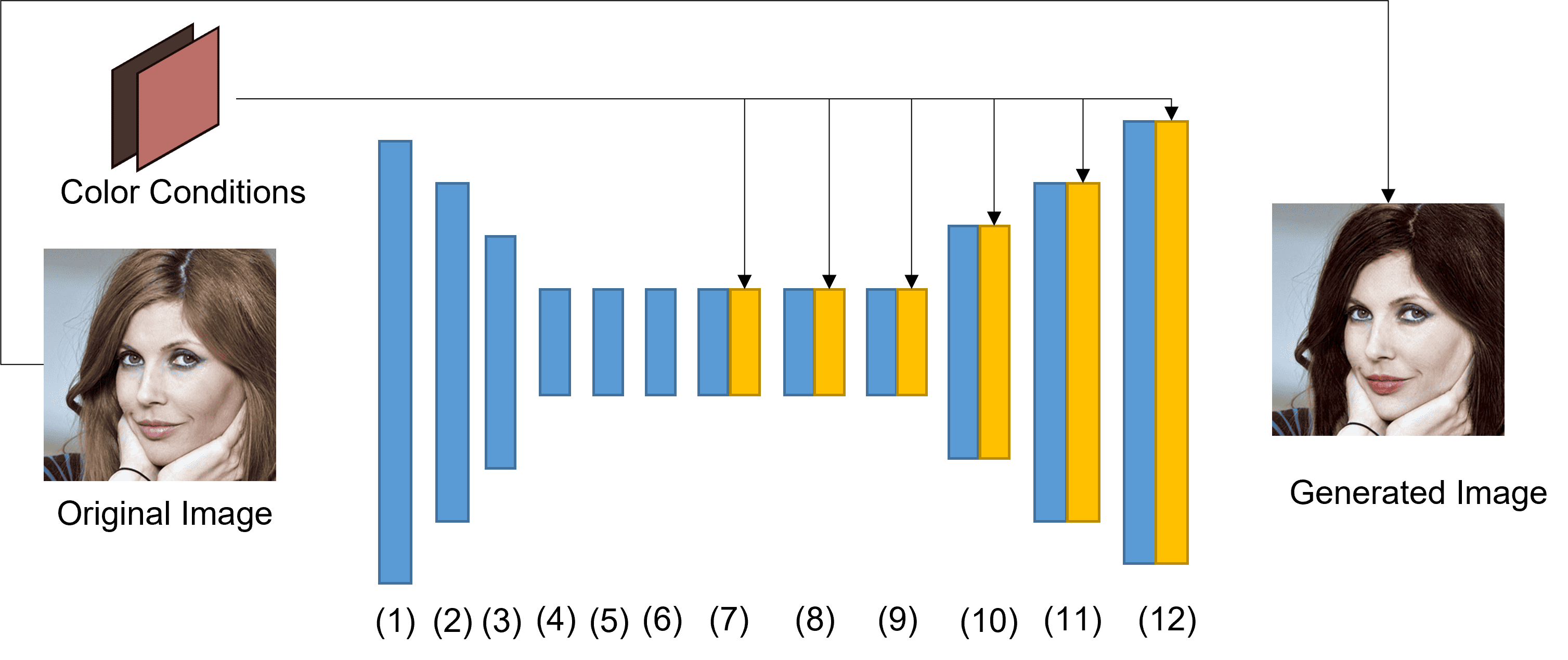}
    \caption{Illustration of conditions incorporation in the generator. The colors are concatenated with the outputs of the convolutional layers. Finally the input image is added to the network's output. Numbers refer to the outputs of the different layers described in table \ref{tab:gen_arch}.}
    \label{fig:tiled_concat}
\end{figure}

\paragraph{Model's architectures}
\label{architectures}
The Generator is a neural network with a common encoder-decoder architecture. It is described in table \ref{tab:gen_arch}. The main input of the network is the image.
The color conditions are fed to the generator using a depthwise concatenation with the different features of the decoder part of the generator. The procedure is described in\cite{DBLP:journals/corr/abs-1711-09020}\cite{liu2019stgan}. The color values are spatially duplicated to match the right size depending on the considered layer. This is illustrated in Figure \ref{fig:tiled_concat}.
The Discriminator network is kept simple with four convolutional and one fully-connected layers. The full details are given in Table \ref{tab:disc_arch}.

\begin{table}
\label{tab:disc_arch}
\centering
\caption{This table details the Discriminator network
architecture.}
\begin{tabular}{|ccc|}
\hline
Discriminator & Activation & Output Shape \\
\hline
Input image & - & 256 x 256 x 3 \\
Conv 3x3 & LReLU & 128 x 128 x 64 \\
Conv 3x3 & LReLU & 64 x 64 x 128 \\
Conv 3x3 & LReLU & 32 x 32 x 256 \\
Conv 3x3 & LReLU & 16 x 16 x 512 \\
Fully-Connected & linear & 1 x 1 x 1 \\
\hline

\end{tabular}
\end{table}

\begin{table}
\label{tab:gen_arch}
\centering
\caption{This table details the Generator network architecture.}
\centering
\begin{tabular}{|ccc|}
\hline
Discriminator & Activation & Output Shape \\
\hline
Input image & - & 256 x 256 x 3 \\
Conv 3x3 (1) & ReLU & 256 x 256 x 64 \\
Conv 3x3 (2) & ReLU & 128 x 128 x 128 \\
Conv 3x3 (3) & ReLU & 64 x 64 x 256 \\
\hline
Residual Block (4) & ReLU & 64 x 64 x 256 \\
Residual Block (5) & ReLU & 64 x 64 x 256 \\
Residual Block (6) & ReLU & 64 x 64 x 256 \\
Residual Block (7) & ReLU & 64 x 64 x 256 \\
Residual Block (8) & ReLU & 64 x 64 x 256 \\
Residual Block (9) & ReLU & 64 x 64 x 256 \\
\hline
DeConv 3x3 (10) & ReLU & 64 x 64 x 256 \\
DeConv 3x3 (11) & ReLU & 128 x 128 x 128 \\
DeConv 3x3 (12) & Tanh & 256 x 256 x 64 \\
Output image & - & 256 x 256 x 3 \\
\hline

\end{tabular}

\end{table}

\subsection{$\lambda_{foc}$ analysis}
\label{annex:l_mask_analysis}
\setlength{\tabcolsep}{4pt}
\begin{table}
\label{tab:mask_study}
\centering
\caption{Metrics comparison for different values of $\lambda_{foc}$. First rows give references values of the $FID$ and $\mathcal{L}^{color}$ between $X_1$ and $X_2$. We find $\lambda_{foc}=600$ as the best trade-off between all the metrics.For all metrics, lower is better. Best results are in bold and second best in italic.}
\begin{tabular}{|l|ccc|}
\hline
Model & $L_{bg}^{avg}\times10^3$ & $FID_{1\rightarrow2}$ & $L_{1\rightarrow2}^{color}$ \\
\hline
Identity & \textbf{0} & \textbf{0.045} & 25.2 \\
$\lambda_{foc}=0$ & 8.00 & 0.060 & \textbf{4.8} \\ 
$\lambda_{foc}=100$ & 4.00 & 0.053 & 5.10 \\ 
$\lambda_{foc}=250$ & 2.64 & 0.052 & \textit{4.95} \\ 
$\lambda_{foc}=600 - ours$ & 0.92 & 0.050 & 5.03 \\ 
$\lambda_{foc}=1000$ & 0.89 & \textit{0.048} & 6.42 \\ 
$\lambda_{foc}=4000$ & 0.87 & 0.051 & 7.05 \\ 
$\lambda_{foc}=6000$ & \textit{0.78} & 0.056 & 9.49 \\
\hline

\end{tabular}

\end{table}
\setlength{\tabcolsep}{1.4pt}
We present here a metrics analysis of the impact of $\lambda_{foc}$ during the training.
We observe in table \ref{tab:mask_study} that optimising the object focus loss fulfills its objective that was limiting the modification of pixels outside of the colored object. Indeed, the higher $\lambda_{foc}$ is, the smaller $L_{bg}^{avg}$ is. This loss also impacts realism and color precision. Lower $\lambda_{foc}$ values improve results, while higher values tend to have a negative influence. Penalizing too much the training through this loss tends to minimize the accepted modifications and does not let the model reach the best color precision possible. A compromise has to be found between these different objectives and for this dataset, we found $\lambda_{foc}=600$ to be the best trade off.

\subsection{Color control using StyleGAN}
\label{annex:stylegan_strategy}
\begin{figure}
    \centering
    \includegraphics[width=0.7\linewidth]{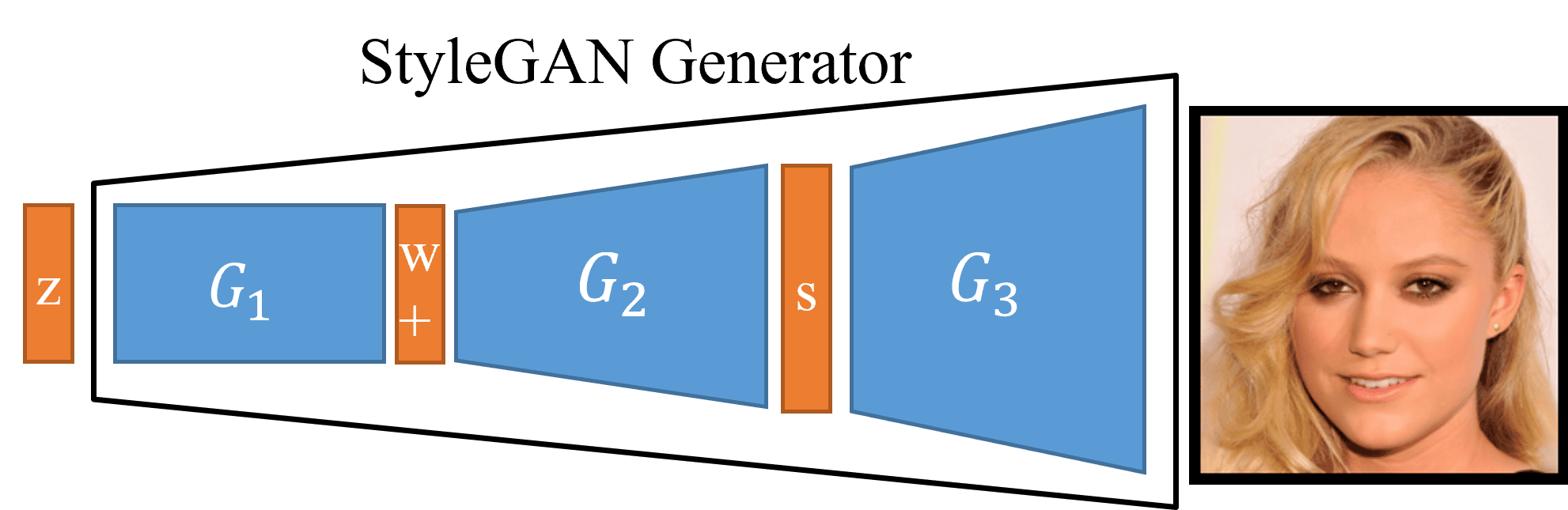}\\
    \vspace{0.8cm}
    \includegraphics[width=1.0\linewidth]{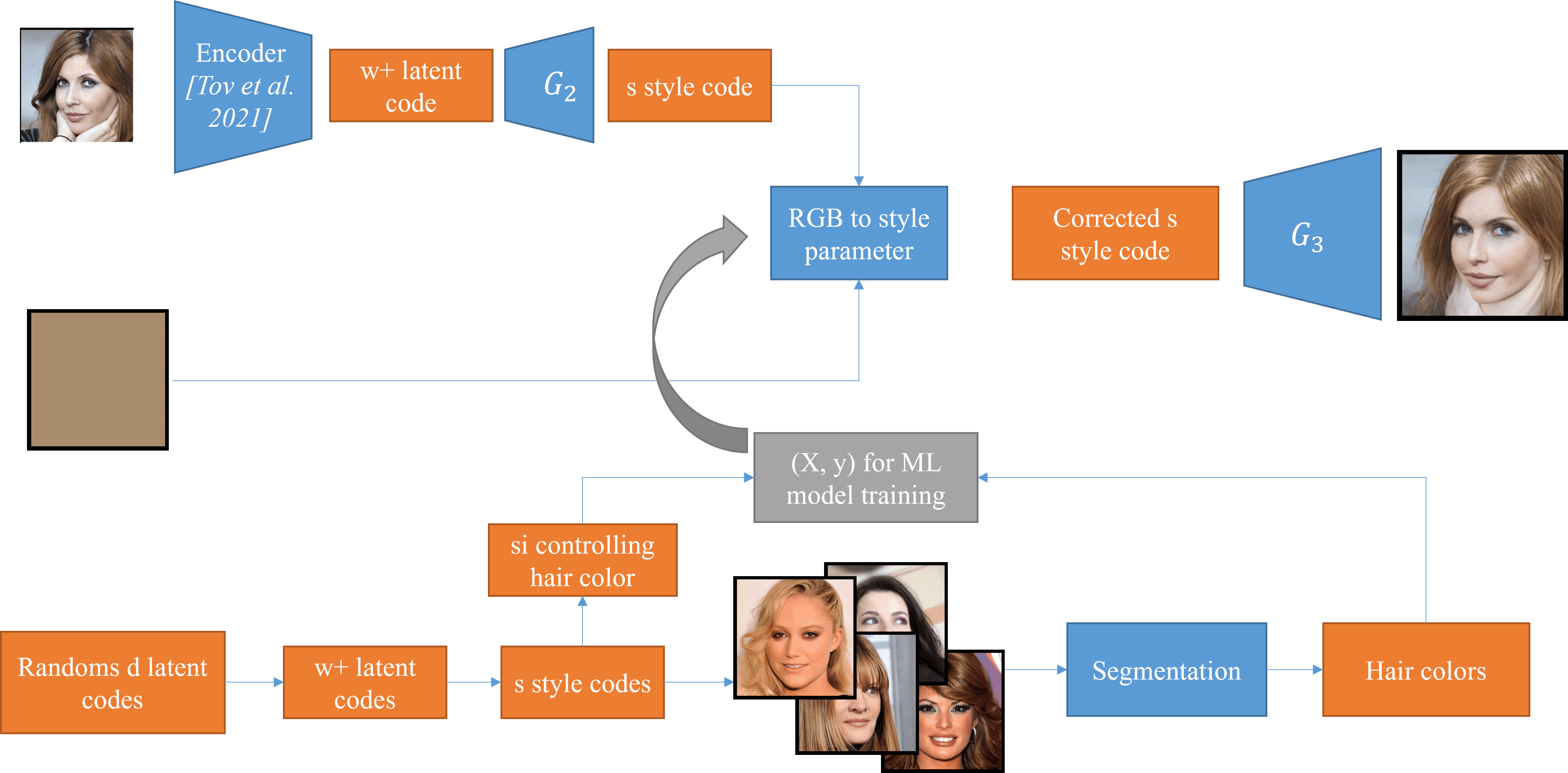}
    \caption{Top graphic represents a simplified decomposition of StyleGAN's model Generator. In particular it shows the two intermediate states that are used, $\mathcal{W+}$ and $\mathcal{S}$. Bottom graphics illustrates the process used to achieve color control with StyleGAN's Generator.}
    \label{fig:stylegan_method}
\end{figure}
StyleGAN's original purpose is to generate realistic images from random vectors. Its effectiveness has led to the development of diverse methods for image to image translation. We wanted a tangible comparison element with our framework, and have investigated StyleGAN. We present here  how to add color control to this existing model. This method can edit an object's color from a given RGB value, like our framework. We have specifically focussed on hair color for this study.

We created a dataset $\mathcal{X}$ of a thousand images randomly generated with StyleGAN. For each $X \in \mathcal{X}$ corresponds a style vector $s\in S$ with $S$ the StyleSpace described in\cite{DBLP:journals/corr/abs-2011-12799}.
In this paper, the auhtors also describe how to only modify $s^j$, specific elements of $s$ as to only modify hair color of the generated image. By choosing the right $ds$ as only to change these $s^j$ elements, we create the two new style vectors $s+ds$ and $s-ds$.  Using StyleGAN's model, this leads to two manipulations of $X$ with different hair colors, $X_{ds^-}$ and $X_{ds^+}$. We are now able to also compute $dc^+=C(X_{ds^+})-C(X)$ and $dc^-=C(X_{ds^-})-C(X)$.
This is enough to have a sufficient dataset to train a machine learning model to predict $ds$ from the desired hair color difference, $dc$.We can now automatically modify a StyleSpace vector as to modify the hair color according to a desired RGB value.

The workflow to use StyleGAN for hair coloring illustrated in Figure \ref{fig:stylegan_method} is as follows:

\begin{itemize}
    \item Encode the original image in StyleGAN's latent space, using \cite{10.1145/3450626.3459838}
    \item This first encoding, $w\in W$ leads to a $s\in S$ StyleSpace vector
    \item Computing $ds$ to reach the desired hair color
    \item Generate resulting image from corrected $s+ds$ using StyleGAN generator
\end{itemize}

The objective here was not to find the best method to control color generations using StyleGAN. It was merely to find a simple workaround that would let us compare our approach with one of the most performing method in image generation.

\subsection{Additional results}

\label{annex:additionnal_results}
\begin{figure}
    \centering
    \includegraphics[width=1.0\linewidth]{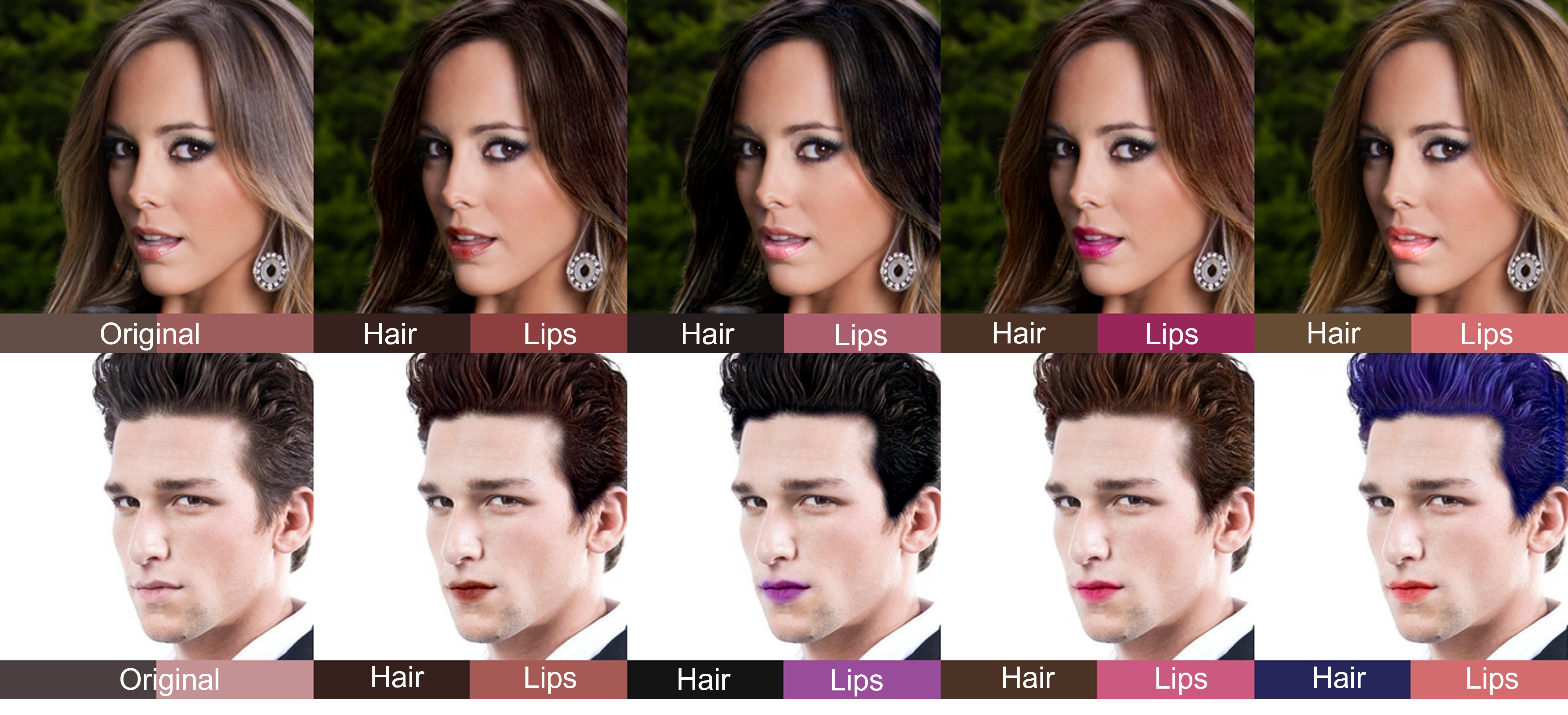}
    \caption{Example results of our framework on hair and lips color edition. Color conditions are indicated under generated images.}
    \label{fig:additional_selfie}
\end{figure}

\begin{figure}
    \centering
    \includegraphics[width=1.0\linewidth]{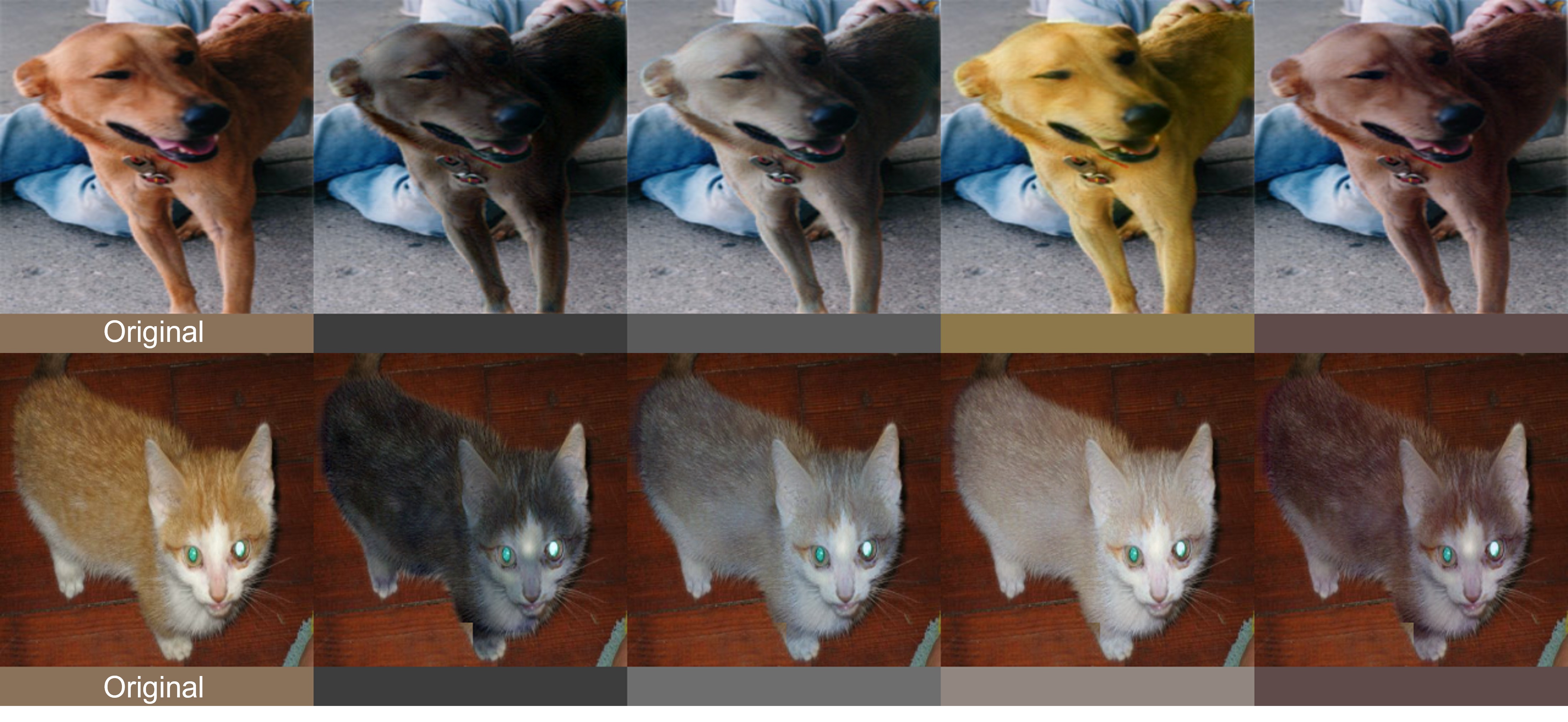}
    \caption{Example results of our framework on cat and dogs color edition. Color conditions are indicated under generated images.}
    \label{fig:additional_catdogs}
\end{figure}

\begin{figure}
    \centering
    \includegraphics[width=1.0\linewidth]{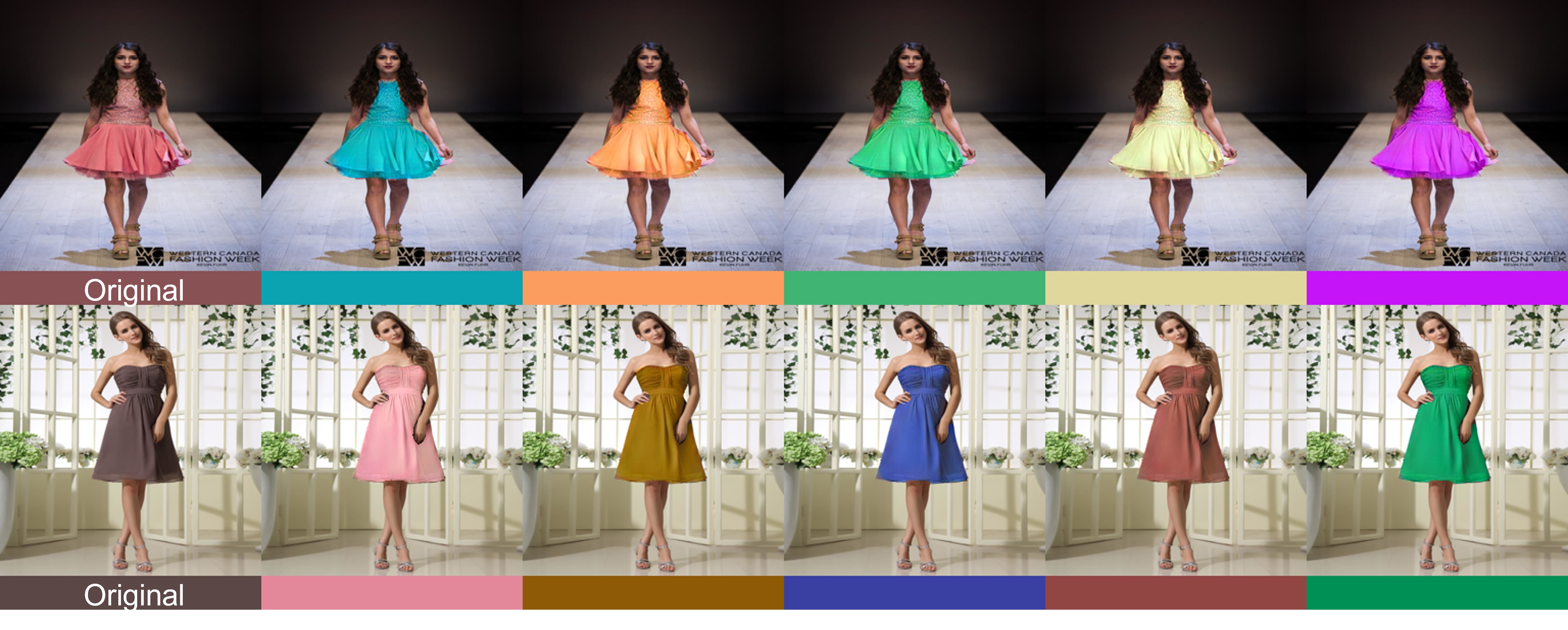}
    \caption{Example results of our framework on clothes color edition. Color conditions are indicated under generated images.}
    \label{fig:additional_clothes}
\end{figure}

\textbf{Color Edition Control}
We present here a few additional results of our model on the three use cases: lips and hair in Figure \ref{fig:additional_selfie}, cats and dogs in Figure \ref{fig:additional_catdogs} and clothes in Figure \ref{fig:additional_clothes}.


If color edition is the application targeted by our framework, it can also be used for color transfer and colorization.

\begin{figure}
    \centering
    \includegraphics[width=0.8\linewidth]{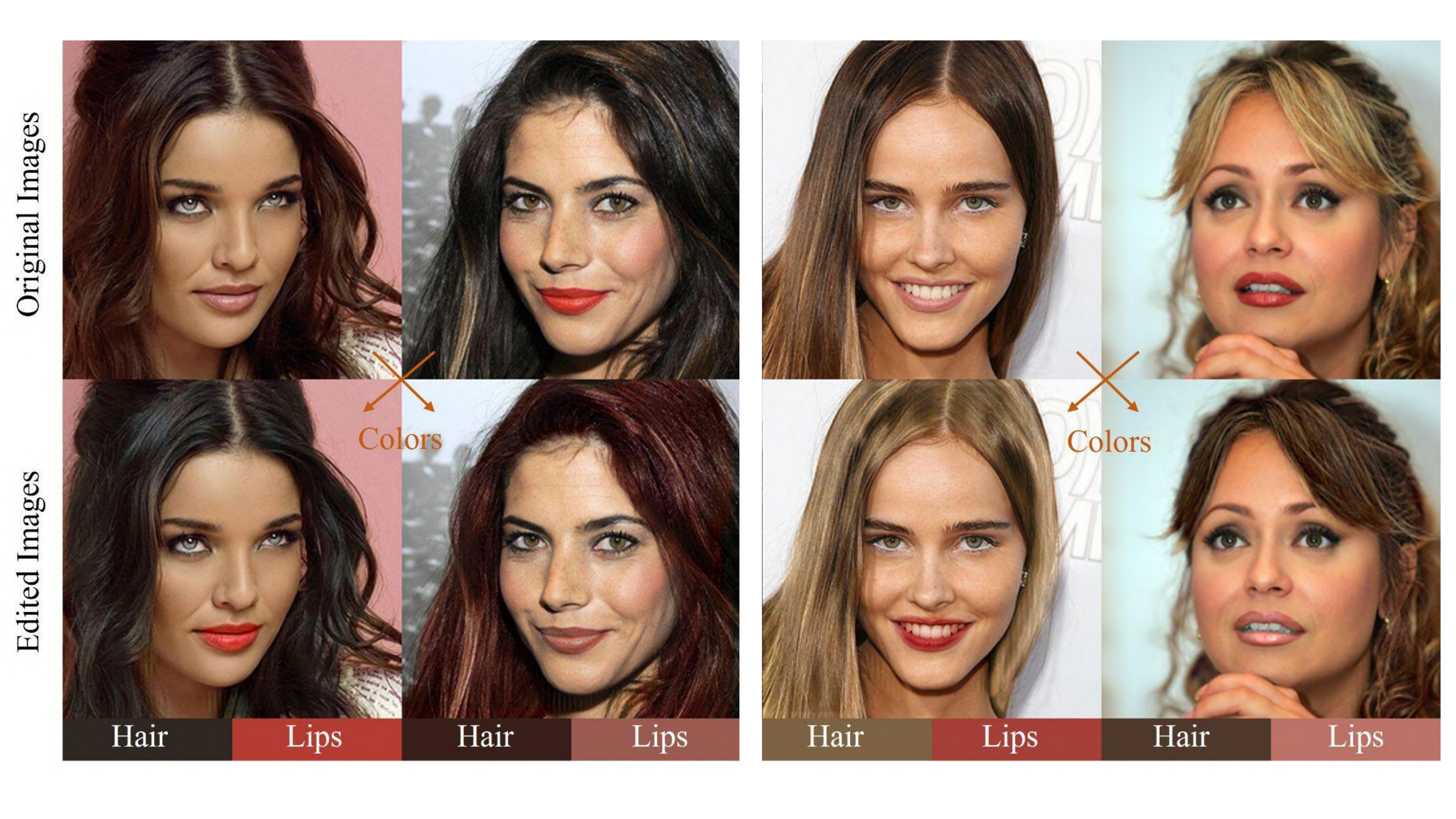}
    \caption{Example of color translation results. First row shows original images and second row shows results of generation using the neighbour's image as the desired colors.}
    \label{fig:color_translation}
\end{figure}

\textbf{Color Transfer}
From a color reference image $X$ and a content image $\Tilde{X}$, the image with the transferred colors is $G(\Tilde{X}, C(X))$. Our framework can be used easily for color transfer, adding a step of color extraction using $C$. Some results can be found in Figure \ref{fig:color_translation}.

\begin{figure}
    \centering
    
  \includegraphics[width=0.8\linewidth]{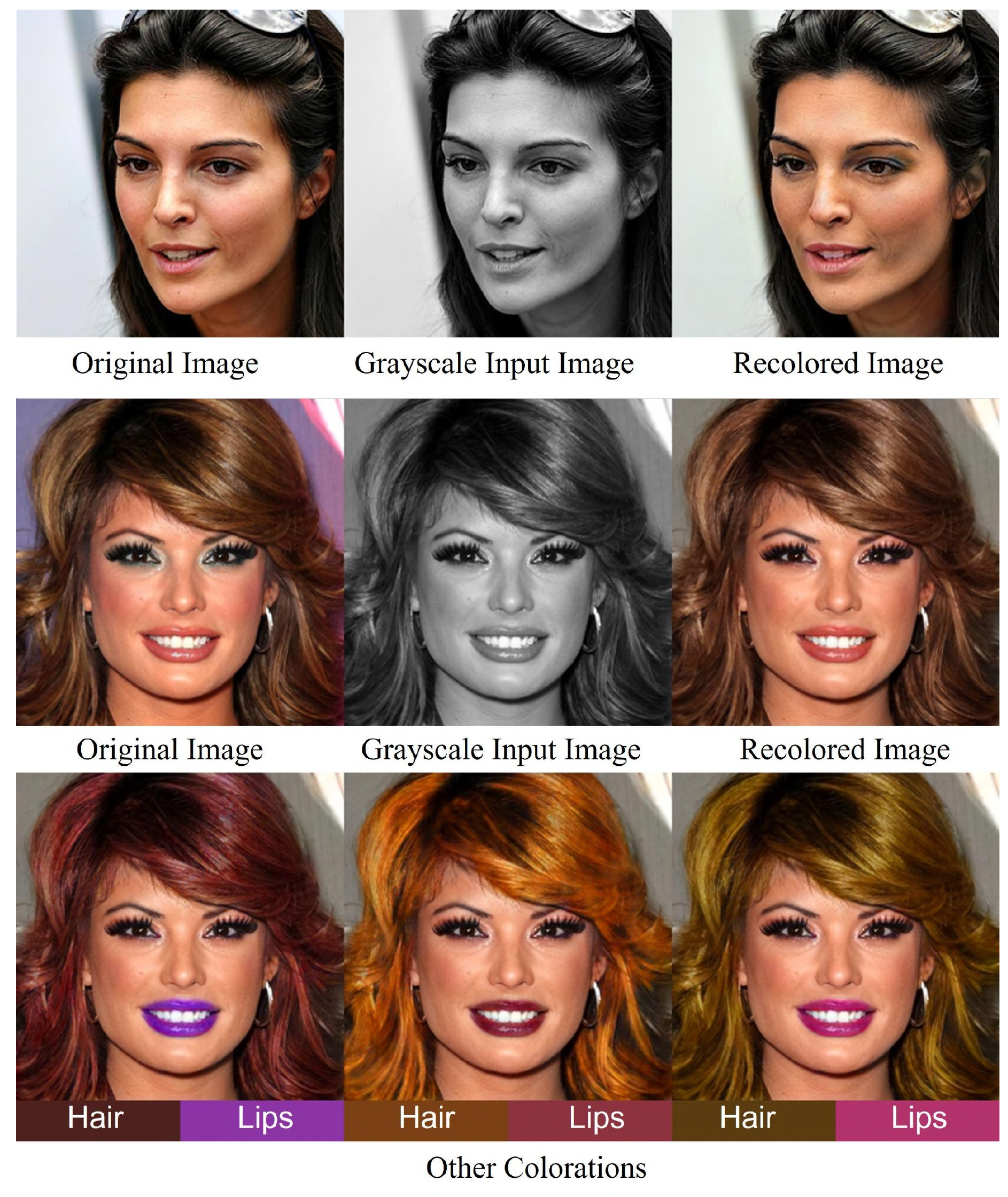}
    \caption{Example results for colorization. The recolored image is generated using the hair and lips color extracted from the original image and given to the model along the grayscale image. On third row is colorization with additional generated colors.}
    \label{fig:gray_rec}
\end{figure}

\begin{figure}[h]
    \centering
    \includegraphics[width=0.8\linewidth]{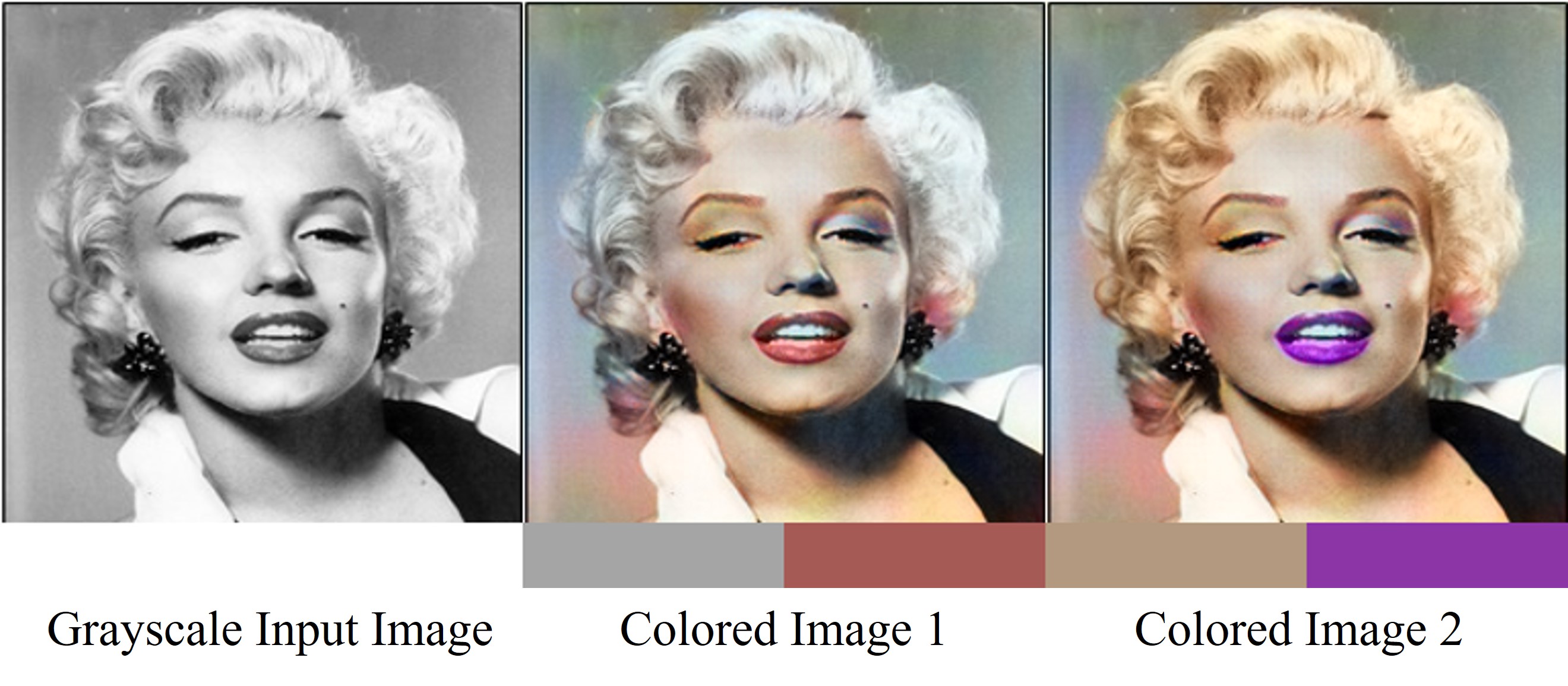}
    \caption{Example results for colorization. Different variations from different hair and lips colors are generated from an originally grayscale image.}
    \label{fig:gray_marilyn}
\end{figure}

\textbf{Colorization}
By feeding $G$ the gray-scale version of the images $X$ during the training, it possible to obtain a colorization model. The benefit being that it is possible to control the future colors for the defined objects. Figure \ref{fig:gray_rec} shows results of gray images recolorization and some possible variations. Figure \ref{fig:gray_marilyn} shows results on an originally gray image. If backgrounds tend to stay in gray colors, it is particularly performing at reconstituting hair and lips colors.


\end{document}